\documentclass{article}
\usepackage{spconf,amsmath,graphicx,hyperref}
\usepackage[T1]{fontenc}
\usepackage{enumitem}

\title{Incremental LSTM-based Dialog State Tracker}
\name{Lukas Zilka, Filip Jurcicek\thanks{This research was partly funded by the Ministry of Education, Youth and Sports of the Czech Republic under the grant agreement LK11221, core research funding, GAUK grant 2076214, SVV project number 260~224 of Charles University in Prague. This work has been using language resources distributed by the LINDAT/CLARIN project of the Ministry of Education, Youth and Sports of the Czech Republic (project LM2010013). Cloud computational resources were provided by the MetaCentrum under the program LM2010005 and the CERIT-SC under the program Centre CERIT Scientific Cloud, part of the Operational Program Research and Development for Innovations, Reg. no. CZ.1.05/3.2.00/08.0144. We gratefully acknowledge the support of NVIDIA Corporation with the donation of the Titan Z GPU used for this research.}}
\address{Charles University in Prague, Faculty of Mathematics and Physics \\ Malostranske namesti 25, 118 00 Prague}

\begin{document}
\maketitle

\begin{abstract}
A dialog state tracker is an important component in modern spoken dialog systems.
We present an incremental dialog state tracker, based on LSTM networks.
It directly uses automatic speech recognition hypotheses to track the state.
We also present the key non-standard aspects of the model that bring its performance close to the state-of-the-art and experimentally analyze their contribution:
including the ASR confidence scores, abstracting scarcely represented values, including transcriptions in the training data, and model averaging.
\end{abstract}
\section{Introduction}
\label{sec:intro}
A dialog state tracker is an important component of statistical spoken dialog systems. It estimates the user's goals throughout the dialog by analyzing the automatic speech recognition (ASR) outputs for the user's utterances. For example, in the restaurant information domain, the dialog state tracker can track what kind of food the user wants and which price range is he looking for, as a probability distribution over \emph{food} and \emph{price\_range}: $\operatorname{P(\mbox{food}, \mbox{price\_range})}$.

The state-of-the-art dialog state trackers~\cite{williams2014web,henderson2014word,lee2014optimizing,smith2014comparative,sun2014sjtu} achieve their top performance by learning from annotated data, and they were shown to work well in the restaurant information domain in the dialog state tracking challenge DSTC2~\cite{henderson2014second}. However, they still possess two undesirable properties. First, they can only track the dialog state turn-by-turn (as opposed to a more complicated word-by-word approach), which limits their interaction with users: For example, in a typical dialog system~\cite{duvsek2014alex,young2013pomdp}, the dialog system can neither provide affirmative natural feedback while the user is speaking, nor can the system interpret additional information said by the user while the system is speaking, both of which is very natural in human-human communication.
And second, some of the trackers use an intermediate semantic representation and a spoken language understanding (SLU) component~\cite{wang2005spoken}. As the representation is manually crafted, it can cause loss of information, and an SLU, if used, is an additional component of the dialog system that needs to be trained and tuned.

The main contribution of this paper is an extension of our LSTM-based~\cite{hochreiter1997long} dialog state tracker, first described in~\cite{zilka2015lectrack}, which brings its performance close to the state-of-the-art models.
We refer to the tracker as LecTrack\footnote{(L)STM R(ec)urrent Neural Network Dialog State (Track)er.}.
LecTrack naturally operates incrementally, word-by-word, and does not require an SLU.
It learns from dialog sessions annotated by dialog state component labels at different time steps.
The improvements consist of including the ASR confidence scores, abstracting scarcely represented values, including transcriptions in the training data, and model averaging.

The paper is organized as follows: First, we give a basic description of the dialog state tracking task in \autoref{sec:dialog-state-tracking}. In \autoref{sec:lstm_dialog_state_tracker}, the model of our LSTM dialog state tracker is described with its training procedure. The tracker is evaluated in \autoref{sec:experiments}. Related work from the literature is discussed in \autoref{sec:related}. \autoref{sec:conclusion} concludes the paper.

\section{Dialog State Tracking}
\label{sec:dialog-state-tracking}
The task of dialog state tracking is to monitor progress in the dialog and provide a compact representation of the dialog history in the form of a \emph{dialog state}~\cite{henderson2014second,zilka2013comparison}. Because of uncertainty in the user input, statistical dialog systems maintain a distribution over all possible states, called the \emph{belief state}. As the dialog progresses, the dialog state tracker updates this distribution given new observations.

In this paper, we define the dialog state at time $t$ as a vector $s_t \in C_1 \times ... \times C_k$ of $k$ dialog state components, sometimes called slots in the literature.
Each component $c_i \in C_i=\{v_1, ..., v_{n_i}\}$ takes one of $n_i$ values, and we assume the components are independent:
$$P(s_t|w_1, ..., w_t)=\prod_i p(c_i|w_1, ..., w_t; \theta)$$
Our dialog state tracker, that we describe in the following, gives the probability distribution only over one of the independent components $p(c_i|w_1, ..., w_t)$. A prediction for more components together is made independently by running different models, specific for each component $i$.

\section{LSTM Dialog State Tracker}
\label{sec:lstm_dialog_state_tracker}
In this section we describe an extended version of the LecTrack LSTM dialog state tracking model~\cite{zilka2015lectrack}. 
The task of the tracker is to map a sequence of words in the dialog to a probability distribution over the values of a dialog state component $p$. For example, for the dialog state component \emph{area}, $p_t$ is a probability distribution over values $\{north, south, east, west\}$ at the time $t$. Because the input words may be preprocessed, we refer to them sometimes as tokens; a sequence of words/tokens $a_1, ..., a_t$ from some a vocabulary $a_i \in Vocab$.

\subsection{Model}
\begin{figure}

\begin{minipage}[b]{1.0\linewidth}
  \centering
  \centerline{\includegraphics[width=8.5cm]{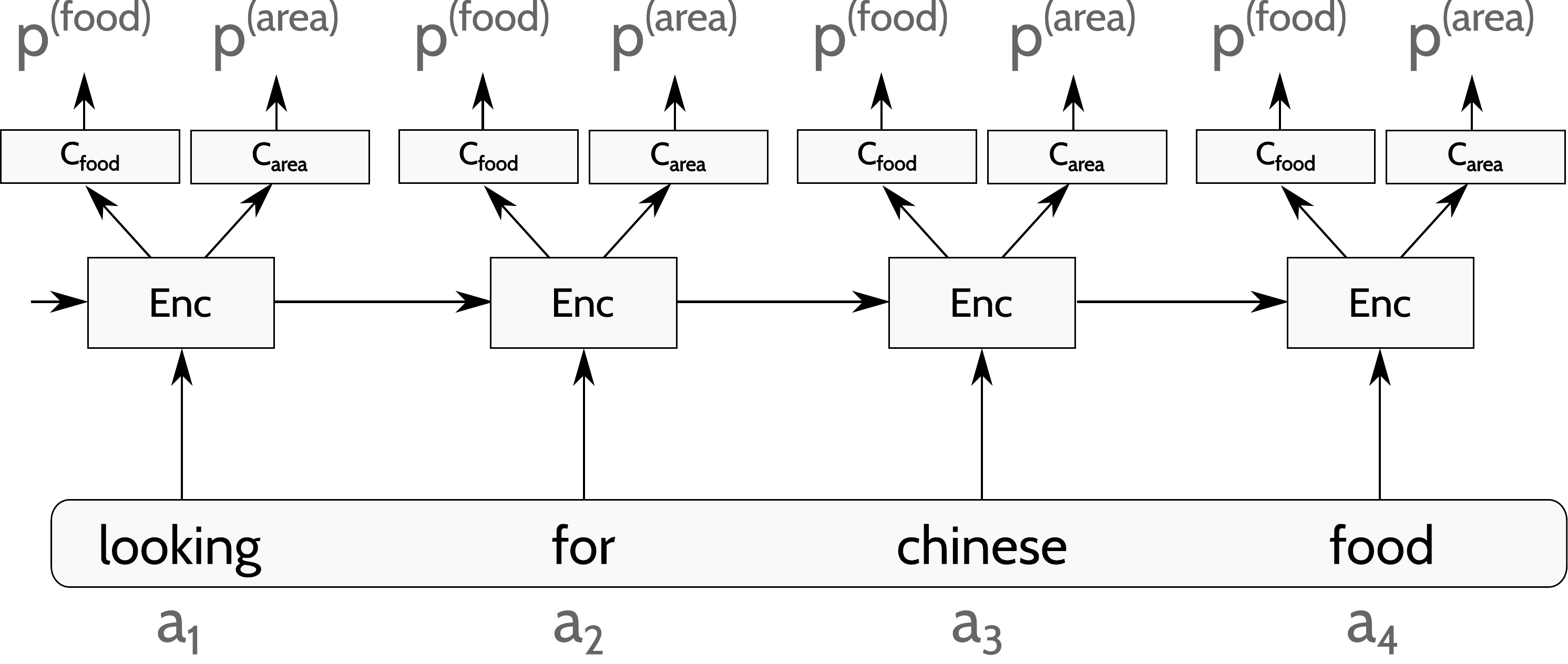}}
\end{minipage}
\caption{A demonstration of the LecTrack LSTM dialog state tracker applied to a user utterance ``looking for chinese food''. The encoding LSTM model $Enc$ is sequentially applied to each input word and its hidden state is used to feed to the state component classifiers.}
\label{fig:lectrack}
\end{figure}

Our dialog state tracking model is an encoder-classifier model: LSTM~\cite{hochreiter1997long}\footnote{Contrary to the original LSTM formulation we use $tanh$ activation instead of $sigmoid$ for the input gate.} is used to encode the information from the input word sequence into a fixed-length vector representation, and given this representation, a classifier returns a probability distribution over the values for the dialog state. The input consists of words which were recognized by ASR along with their confidence scores. The words are represented as embeddings, and before they are passed to the LSTM, a single-layer neural network is used to create new word embeddings, accounting for the confidence score. An example of the model applied to a particular input sentence is at~\autoref{fig:lectrack}.

Formally, we have an input neural network that maps the word $a$ and its ASR confidence score $r$ to a joint representation $u$:
$$u=\operatorname{NN}(a, r)$$
The representation $u$ is used by the LSTM encoder along with the previous hidden state $q_{t-1}=(c_{t-1}, h_{t-1})$\footnote{The state of a standard LSTM model consists of two components.} to create a new hidden state $q_{t}$:
$$q_t=\operatorname{Enc}(u, q_{t-1})$$
The classifier, represented by a single softmax layer, then maps the hidden state to a probability distribution over all possible values:
$$p_t=\operatorname{C}(h_t)$$

Put together, these components make up LecTrack, which maps an input ASR word and score sequence into a sequence of dialog state estimates
\begin{align*}
\bf{LecTrack}: & ~(a_1, r_1) ..., (a_n, r_n) \rightarrow p_{1}, ..., p_{n} \\
\forall i \in {1, ..., n}: & ~q_i = (c_i, h_i) = \operatorname{Enc}(\operatorname{NN}(a_1, r_1), q_{i-1}) \\
\forall i \in {1, ..., k}: & ~p_i = \operatorname{C}(h_i)
\end{align*}
where $n$ is the length of the input sequence.

\subsection{Improvements}
\label{sec:improve}
In this subsection, we describe the modifications we make to the original tracker~\cite{zilka2015lectrack}.

Note that contrary to the original model, the new model described in this paper is factored by dialog state components.
We empirically found that it converges faster and more reliably, and it also makes the model simpler and standard (because now the model is a standard multi-class classification, as opposed to a multi-target classification before, which allows the use of the standard neural-network toolkits available on Internet for implementation).
Tracking of more dialog state components is achieved simply by instantiating more LSTM models in parallel.

\subsubsection{Including ASR scores}
The original model did not make use of the ASR 1-best hypothesis confidence scores.
When the model does not have this information, the only possible way to learn not to trust the input is by learning which word patterns correspond to well-recognized speech and which are typical for erroneously-recognized speech.
Therefore, we decided to include the confidence score of the input hypothesis as an additional dimension to each input word embedding, and add one fully-connected non-linear layer between this input and the LSTM, so that the model can learn to transform the embeddings according to the confidence score.

\subsubsection{Including Transcriptions in Training Data}
Our training data are noisy due to ASR errors, and it is a common practice to expand the training set to reduce the noise.
We thus decided to mix the ASR 1-best hypotheses with the true manually-transcribed user utterances to form an expanded training set, which should reduce the amount of noise.

\subsubsection{Model Averaging}
Following~\cite{williams2014web,henderson2014second}, where the authors successfully use a simple model averaging strategy to boost the performance of their models, we train 10 different models from 10 different random initializations and average their predictions.

\subsubsection{Abstracting low-occurring values}
Our model has little chance of learning to properly predict state component values that do not occur frequently in the training data set.
We thus decided to substitute the ones that occur less than 40 times in the training set by an abstract value.
As a result, we replace each occurrence of such a~low-occurring value by an abstract token, e.g. jamaican for \emph{\#food1}. Occurrences of the same value are replaced by the same abstract token, and if a different value is encountered we create another abstract token, e.g. \emph{\#food2}. For each of these abstract tokens we need to add a new class to the classifier.
During tracking, the classifier output is post-processed and these values are substituted back, e.g. prediction \emph{\#food1} for \emph{jamaican}.

This modification makes the tracker able to track values that it has never seen in the training data, by manually putting them in the abstraction dictionary.
Also, because the frequent values are not being abstracted the tracker can still learn ASR error patterns for them.
This idea is similar to~\cite{henderson2014second} who abstracted out everything and included both abstracted and non-abstracted features as the input to his model.

\subsection{Training}
The training criterion is a cross-entropy loss~\cite{rubinstein2004cross} for a dialog example, which is annotated by true lables at some points in time:

$$ l(\theta) = -\sum_{t \in Y} log \operatorname{LecTrack}(a_1, r_1, ..., a_n, r_n)^{t}_{y_t}$$

Here, $y_i$ denotes a label for the dialog state at time $i$, and $Y$ is a set of times where the label $y_i$ exists (times that correspond to the end of turns, because in our experiments we have labels only for them). $\operatorname{LecTrack}(.)^m_n$ denotes the probability of the $n$-th value at time $m$.

We fit the model using ADAM optimization algorithm~\cite{kingma_adam:_2014}. All parameters are initialized randomly from a zero-mean Gaussian with $\sqrt{\frac{2.0}{d}}$ variance~\cite{sussillo14} (where $d$ is the dimensionality of the input for the layer), apart from the biases of the LSTM forget gates, which are initialized to to $1.0$~\cite{jozefowicz2015}.

After each optimization epoch, we monitor the performance\footnote{See~\autoref{sec:metrics} for the description of the featured metrics.} of the model on a held-out set $D$. When the performance stops increasing for several iterations, we terminate the training and select the best-performing model.

\section{Experiments}
\label{sec:experiments}
\subsection{Dataset}
To train and evaluate our model, we use the DSTC2~\cite{henderson2014second} data set. The DSTC2 data consists of about 3,000 dialogs from the restaurant information domain, each dialog is 10 turns long on average. The data is split into training, development and test sets.

Our model is incremental and does not explicitly represent turns, but the DSTC2 data set contains only turn-based dialogs. So we treat each dialog in DSTC2 as a sequence of words in time, where the dialog state labels are always attached to the last word of the turn. Ideally we would run the evaluation on a data set where we could also measure the incremental capabilities of the tracker, but to the best of our knowledge, no such data set is publicly available yet, and we leave the collection and experimentation on such a data set for future work.

In our experiments, the word embeddings have 170 dimensions, the input network has 300 output units with ReLU on top, LSTM encoder has 100 cells, and we train using full network unrolling in time in mini-batches of 10 dialogs.

\subsection{Baseline}
A baseline system for this domain has been provided by the DSTC2 organizers. It uses the SLU results and confidence to rank hypotheses for the values of the individual dialog state components. There were several baselines described in~\cite{henderson2014second}; we report the results of the \emph{focus} baseline, which was the best among them.

\subsection{Data Preprocessing}
Each dialog turn consists of the system and the user utterance. We serialize both of them into a stream of pairs \texttt{(word/token, ASR confidence score)} as the input to our model.

\paragraph*{System Utterance Preprocessing:}
To get the the system input, we perform a simple preprocessing. We flatten the system dialog acts of the form \texttt{act\_type(slot\_name=slot\_value)} into a sequence of three tokens $t_1, t_2, t_3$, where $t_1=\operatorname{act\_type}$, $t_2=\operatorname{slot\_name}$ and $t_2=\operatorname{slot\_value}$. For example \texttt{request} \texttt{(slot=food)} is converted into $request, slot, food$, which the model then sees as a word sequence of length three.

\paragraph*{User Utterance Preprocessing:}
For the sake of simplicity, we use only the best live-ASR hypothesis\footnote{There are batch and live ASR results in the DSTC2 data. We use the live ones and refer to them as live-ASR.} (we refer to it as ASR 1-best) and ignore the rest of the n-best list.
It is not obvious how to incorporate more ASR hypotheses into an incremental dialog state tracker in a good way and we plan to address this issue in our future work.

\paragraph*{Out-of-Vocabulary Words:}
Out-of-Vocabulary words are randomly mixed into the training data to give the model a chance to cope with unseen words: At training time, a word in the user input is replaced by a special out-of-vocabulary token with a probability $\alpha$\footnote{Throughout this paper we use $\alpha=0.1$.}. At test time, this token is used to represent all unknown words.

\subsection{Evaluation Metrics}  
\label{sec:metrics}
We follow the DSTC2 methodology~\cite{henderson2014second} and measure the accuracy and L2 norm of the joint slot predictions.
The joint predictions are grouped into the following groups: Goals, Requested, Method. The results of each group are reported separately.

For each dialog state component in each dialog, the measurements are taken \emph{at the end of each dialog turn}\footnote{The measurements are taken at the end of each dialog turn, provided the component has already been mentioned in some of the SLU n-best lists in the dialog. Note we do not use the SLU n-best list in our model at all, but we adapt this metric to be able to compare to the other trackers in DSTC2.}.

To asses the effect of the individual improvements over the base model described in~\autoref{sec:improve},
we evaluate the following configurations that cumulatively add the different improvements on top of each other:
\begin{description}[labelwidth=0em,leftmargin=5.5em]
    \item[(base)] Base model without the proposed improvements~\cite{zilka2015lectrack}.
    \item[(score)] Include scores.
    \item[(transcr)] Include transcriptions.
    \item[(abstract)] Abstract low-occurring classes.
    \item[(model avg)] Model Averaging.
    \item[(dontcare oracle)] Don't care oracle (detailed later).
\end{description}

\subsection{Results}

\begin{table*}
    \centering
    \begin{tabular}{|l|llllll|llllll|}
    \hline
                                   & \multicolumn{6}{c|}{Development set}                                                                       & \multicolumn{6}{c|}{Test set}  \\
                                   & \multicolumn{2}{c}{Goal}    & \multicolumn{2}{c}{Method}     & \multicolumn{2}{c|}{Requested}  & \multicolumn{2}{c}{Goal}   & \multicolumn{2}{c}{Method}     & \multicolumn{2}{c|}{Requested} \\
    \multicolumn{1}{|c|}{model}    & Acc.      & L2              & Acc.           & L2            & Acc.        & L2                & Acc.         & L2          & Acc.       & L2                & Acc.          & L2 \\ \hline \hline
    baseline                       & 0.61      & 0.63            & 0.83           & 0.27          & 0.89        & 0.17              & 0.72         & 0.46        & 0.90       & 0.16              & 0.88          & 0.20    \\ \hline
    LecTrack (base)                & 0.63      & 0.74            & 0.90           & 0.19          & 0.96        & 0.08              & 0.62         & 0.75        & 0.92       & 0.15              & 0.96          & 0.07    \\
    LecTrack (score)               & 0.63      & 0.73            & 0.89           & 0.20          & 0.96        & 0.07              & 0.64         & 0.73        & 0.92       & 0.16              & 0.96          & 0.07    \\
    LecTrack (transcr)             & 0.66      & 0.69            & 0.90           & 0.20          & \bf{0.97}   & 0.07              & 0.67         & 0.65        & 0.92       & 0.15              & 0.97          & 0.07    \\
    LecTrack (abstract)            & 0.67      & 0.65            & 0.90           & 0.20          & \bf{0.97}   & 0.07              & 0.68         & 0.64        & 0.93       & 0.14              & 0.97          & 0.06    \\
    LecTrack (model avg)           & 0.69      & 0.71            & 0.90           & 0.19          & \bf{0.97}   & 0.07              & 0.72         & 0.64        & 0.93       & 0.14              & 0.97          & 0.06    \\ \hline
    LecTrack (dontcare oracle)     &           &                 &                &               &             &                   & 0.75         & 0.50        &            &                   &               &         \\ \hline
    turn-based RNN
    \cite{henderson2014word}       & 0.70      & \bf{0.46}       & \bf{0.92}      & 0.14          & \bf{0.97}   &     0.06          &     0.77     & \bf{0.35}   &     0.94   &     0.10     & \bf{0.98}     & \bf{0.04}   \\
    state-of-the-art~
    \cite{williams2014web}         & \bf{0.71} & 0.74            & 0.91           & \bf{0.13}     & \bf{0.97}   & \bf{0.05}         & \bf{0.78}    & \bf{0.35}   & \bf{0.95}  & \bf{0.08}    & \bf{0.98}     & \bf{0.04}   \\ \hline
    \end{tabular}
    \medskip
    \caption{Performance on the DSTC2 data.}
    \label{table:results}
\end{table*}

The results of all evaluated LecTrack configurations on the DSTC2 data are summarized in~\autoref{table:results}.
The results from DSTC2 are publicly available along with the output of the trackers on test data set so we try to compare our tracker to~\cite{henderson2014word}, which we refer to as RNNTrack, to see in greater detail where are our strengths and weaknesses.

LecTrack's accuracy in its strongest configuration (model~avg) is better than the baseline and comes close to the state-of-the-art, with the exception of the \emph{Goal} group on the test data set.
Note that the model has never seen test data set during training, and development data set was used for selecting the best model seen during the training.

\subsubsection{Test set performance difference}
We attribute the performance difference for the Goal group between the development and test data sets to a substantial difference between the dialog systems used for the two data sets.
The training and development data sets were collected using a different dialog system than the test data set.

The dialog system in the test dataset produces on average about 25\% longer system utterances (measured in the number of input words/tokens), which can influence the stability of the LSTM predictions due to increased number of time steps.
Also, the distributions of the slot values differ substantially, particularly for the \texttt{dont care} value.

Other trackers from the literature do not have this issue because they all extract features from the complete turn and thus are not influenced by the length of the utterances.

We believe we can address the different lengths of the system utterances by considering the system utterance separately and injecting it into the stream of user words just as a single special token. This way the system input is always long one token, regardless of the dialog system used.

\subsubsection{Improvements}
Our improvements mostly affect the performance of the tracker on the Goal group.
The base tracker already performs well on the Method and Requested groups so the improvements there are modest.

Model averaging proved to provide a substantial improvement in performance.
This is in accordance with other approaches that also combine multiple models to produce the final predictions.
There is a body of work on compressing the model ensemble back into a single model~\cite{romero2014fitnets,bucilua2006model,ba2014deep}, which appears as an interesting future research direction.

Including true transcriptions in the training yields almost as big improvement to the results as model averaging.
The size of the training corpus is quite small (only about 1500 dialogs) and without the transcriptions, some values were never seen in their correct form in the training data.
Moreover, with the transcriptions, the tracker is more biased towards learning the correct generalization patterns and can learn to correct some typical ASR mistakes.

Including ASR confidence scores only gives a modest performance improvement on the test set.
This fact is surprising because we believe the ASR confidence scores are very important for tracking, otherwise the tracker does not have a clear signal about the correctness of the user utterance.

\subsubsection{Slot \emph{Food}}
The slot \emph{food} is arguably the most difficult slot for the tracker because it takes 91 values and is frequently talked about. Therefore, we chose it for a more detailed analysis of the tracker's results.

The most frequent value in the slot \emph{food} is \texttt{dont care} value, which is the result of the user saying ``I don't care'' after the system prompted him for the type of food he wants.
However, the decision whether the user does not care about food or something else is dependent on the system prompt, and the tracker must make use of this information.
Our system achieves 81\% accuracy for the \texttt{dont care} value, whereas RNNTrack~\cite{henderson2014word} achieves 91\%.
This suggests that our tracker is not able to properly learn the dependency between the system and user utterances.
A brief manual examination of other prediction errors confirms this.

The \texttt{dont care} values makes up 25\% of the correct labels in the test data but only 15\% in the development data, which is another reason for the difference in performance between the two datasets.
Indeed, when we treat RNNTrack as an oracle to provide the \texttt{dont care} and \texttt{null} predictions (for all slots, not just slot \emph{food}), we beat the baseline and come close to state-of-the art (LecTrack (dontcare oracle) line in~\autoref{table:results}).

\section{Related Work}
\label{sec:related}

The only other incremental dialog tracker know to us is used in~\cite{skantze2009incremental}.
In this paper, the authors describe an incremental dialog system for number dictation as a specific instance of their incremental dialog processing framework.
To track the dialog state, they use a discourse modeling system which keeps track of confidence scores from semantic parses of the input;
these are produced by a grammar-based semantic interpreter with a hand-coded context-free grammar.
Unlike our system, it requires handcrafted grammar and an explicit semantic representation of the input.

Using RNN for dialog state tracking has been proposed before~\cite{henderson2014word,henderson2013deep}.
The dialog state tracker in~\cite{henderson2014word} uses an RNN, with a very elaborate architecture, to track the dialog state turn-by-turn.
Similarly to our model, their model does not need an explicit semantic representation of the input.
They also use a similar abstraction of low-occurring values (they call the technique ``tagged n-gram features''), which should result in better generalization on rare but well-recognized values.

We use only 1-best ASR hypothesis and achieve near state-of-the-art results, while the other tracking models from the literature~\cite{williams2014web,henderson2014word,lee2014optimizing,sun2014sjtu} typically use the whole ASR/SLU n-best list as an input.

\section{Conclusion}
\label{sec:conclusion}
We presented new improvements in our LecTrack incremental LSTM-based dialog state tracker~\cite{zilka2015lectrack}, which make the tracker close to state-of-the-art results on the DSTC2 data set.
The tracker works incrementally word-by-word and does not need a separate SLU component.
The largest improvement was achieved by including the transcriptions in the training data set, and by using an ensemble of models.
Minor improvements were brought by including ASR hypothesis scores and value abstraction.
We also demonstrated that it is enough to use 1-best hypothesis only to achieve near state-of-the-art results in dialog state tracking on DSTC2 data set.

In future, we would like to investigate why the ASR hypothesis confidence score does not play a bigger role in our model, what techniques to employ to reduce the need for the model averaging, and how to use the tracker in a real incremental dialog system.

\bibliographystyle{IEEEbib}
\bibliography{strings,refs}

\end{document}